\documentclass{article} 
\usepackage{iclr2025_conference,times}

\usepackage{amsmath,amsfonts,bm}

\def\eqref#1{equation~\ref{#1}}

\def\1{\bm{1}}

\DeclareMathAlphabet{\mathsfit}{\encodingdefault}{\sfdefault}{m}{sl}
\SetMathAlphabet{\mathsfit}{bold}{\encodingdefault}{\sfdefault}{bx}{n}

\usepackage{hyperref}
\usepackage{url}

\usepackage{microtype}
\usepackage{inconsolata}
\usepackage{amsmath}
\usepackage{amsfonts}
\usepackage{amssymb}
\usepackage{subfigure}
\usepackage{cases}
\usepackage{dsfont}
\usepackage{bm}
\usepackage{bbm}
\usepackage{graphicx}
\usepackage{color}
\usepackage{multicol}
\usepackage{multirow}
\usepackage{wrapfig,lipsum,booktabs}
\usepackage[ruled,vlined,linesnumbered]{algorithm2e}
\usepackage{pgfplots}
\pgfplotsset{compat=1.12} 
\usepackage{filecontents}
\usepackage{tikz}
\usepackage{xcolor}
\usepackage{colortbl}
\usepackage{xspace}
\usepackage{makecell}
\usepackage{soul}
\usepackage{subcaption}
\usetikzlibrary{calc}
\usepgfplotslibrary{groupplots}
\usetikzlibrary{angles,quotes} 
\usetikzlibrary{shapes,arrows}
\usetikzlibrary{matrix}
\usetikzlibrary{patterns}
\usepackage{tikz-3dplot}
\usepackage{hyperref}
\usepackage{cleveref}
\usepackage{paralist}
\usepackage{cancel}
\usepackage{todonotes}
\usepackage{tabu}
\usepackage{rotating}
\usepackage{etoolbox}
\usepackage{adjustbox}
\usepackage{enumerate}
\usepackage{enumitem}
\setitemize{itemsep=0pt,topsep=0pt,parsep=0pt,partopsep=0pt}
\setenumerate{itemsep=0pt,topsep=0pt,parsep=0pt,partopsep=0pt}
\usepackage{pifont}
\usepackage{cancel}
\usepackage{lipsum}
\usepackage{listings,lstautogobble}
\usepackage{fancyvrb}
\usepackage{fvextra}
\usepackage{caption}
\usepackage{arydshln}
\usepackage{float}
\usepackage{color} 
\usepackage{colortbl}
\usepackage{pmboxdraw} 
\usepackage{wasysym}
\usepackage{csquotes}
\usepackage{mfirstuc}
\usepackage{fontawesome}

\SetKwComment{Comment}{$\triangleright$\ }{}

\usepackage[most,skins,theorems]{tcolorbox}
\tcbset{
  aibox/.style={
    width=\linewidth,
    top=8pt,
    bottom=4pt,
    colback=blue!6!white,
    colframe=black,
    colbacktitle=black,
    enhanced,
    center,
    attach boxed title to top left={yshift=-0.1in,xshift=0.15in},
    boxed title style={boxrule=0pt,colframe=white,},
  }
}

\newtcolorbox{AIbox}[2][]{aibox,title=#2,#1}

\definecolor{mytheoremfr}{RGB}{122, 106, 226} 
\definecolor{mytheorembg}{RGB}{224, 214, 250} 
\newtcbtheorem{Prompt}{Prompt}{colback=mytheorembg!25,colframe=mytheoremfr, boxrule=0.5pt
}{th}

\newcommand{\model}[1]{\textsc{#1}\xspace}
\newcommand{\ours}{\model{HomerAgents}}
\newcommand{\oursold}{\model{HomerAgents+}}
\newcommand{\oursnew}{\model{HomerAgents-Neo}}

\newcommand{\dataset}[1]{\texttt{#1}\xspace}
\newcommand{\datasetall}{\dataset{OdysseyBench}}
\newcommand{\datasetold}{\dataset{OdysseyBench+}}
\newcommand{\datasetnew}{\dataset{OdysseyBench-Neo}}
\newcommand{\officebench}{\dataset{OfficeBench}}

\newcommand{\temp}[1]{#1\xspace}
\newcommand{\intent}{\temp{task intent}}

\title{OdysseyBench: Evaluating LLM Agents on Long-Horizon Complex Office Application Workflows}

\author{%
Weixuan Wang\textsuperscript{1} $\dagger$ $\heartsuit $  \quad Dongge Han\textsuperscript{2}$\dagger$  \quad Daniel Madrigal Diaz \textsuperscript{2} \quad Jin Xu \textsuperscript{2} \\
\textbf{Victor Rühle} \textsuperscript{2} \quad \textbf{Saravan Rajmohan} \textsuperscript{2}
 \\[1ex]
\textsuperscript{1}School of Informatics, University of Edinburgh  \quad
\textsuperscript{2}Microsoft \\
\texttt{weixuan.wang@ed.ac.uk}   \quad
}
\begingroup
\footnotetext{
\textsuperscript{$\dagger$}~Equal contribution.
}
\footnotetext{
\textsuperscript{$\heartsuit $}~This work was done during an internship at Microsoft.
}
\endgroup

\iclrfinalcopy 
\begin{document}

\maketitle

\begin{abstract}
Autonomous agents powered by large language models (LLMs) are increasingly deployed in real-world applications requiring complex, long-horizon workflows. However, existing benchmarks predominantly focus on atomic tasks that are self-contained and independent, failing to capture the long-term contextual dependencies and multi-interaction coordination required in realistic scenarios. To address this gap, we introduce \datasetall, a comprehensive benchmark for evaluating LLM agents on long-horizon workflows across diverse office applications including Word, Excel, PDF, Email, and Calendar. Our benchmark comprises two complementary splits: \datasetold with 300 tasks derived from real-world use cases, and \datasetnew with 302 newly synthesized complex tasks. Each task requires agent to identify essential information from long-horizon interaction histories and perform multi-step reasoning across various applications. To enable scalable benchmark creation, we propose \ours, a multi-agent framework that automates the generation of long-horizon workflow benchmarks through systematic environment exploration, task generation, and dialogue synthesis. Our extensive evaluation demonstrates that \datasetall effectively challenges state-of-the-art LLM agents, providing more accurate assessment of their capabilities in complex, real-world contexts compared to existing atomic task benchmarks. We believe that \datasetall will serve as a valuable resource for advancing the development and evaluation of LLM agents in real-world productivity scenarios. In addition, we release \datasetall and \ours to foster research along this line.\footnote{\url{https://github.com/microsoft/OdysseyBench.git}}
\end{abstract}

\section{Introduction}

Autonomous agents powered by large language models (LLMs) have demonstrated remarkable capabilities across diverse domains, including reasoning \citep{lin2024lean,boisvert2024workarena++,yao2024tau,wang2024rethinking}, software development \citep{qian2023chatdev,yang2024swe,murty2024nnetnav,zhou2023webarena,xie2025agentsynth}, and scientific research \citep{drouin2024workarena,wu2025agentic,zheng2025deepresearcher}. As these agents increasingly transition from research settings to real-world applications, they are expected to handle complex, multi-step tasks such as drafting professional emails, updating documents, and managing personal calendars \citep{yao2024tau,wang2024officebench,xu2024theagentcompany}. This shift underscores the need for the development of comprehensive benchmarks that accurately reflect real-world scenarios and rigorously evaluate agent performance in complex, contextual task environments.

\begin{figure}[t]
\centering          
\subfigure[Atomic Tasks]{\label{fig:example-others}\includegraphics[scale=0.32]{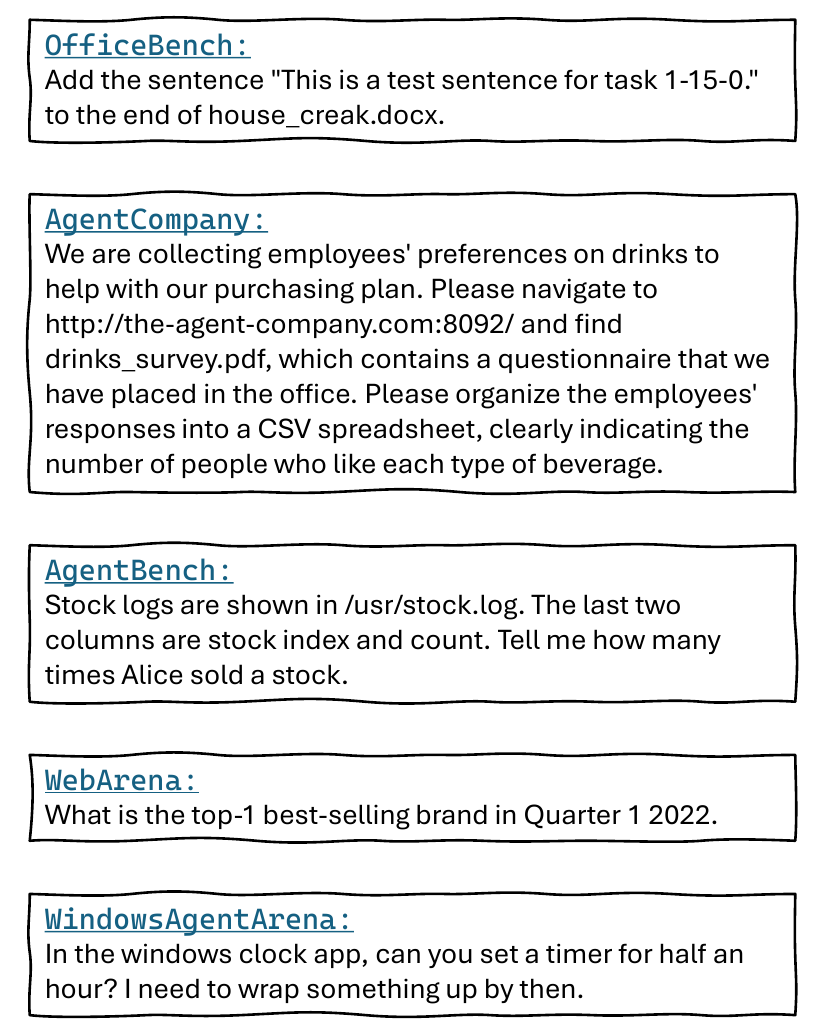}} 
\subfigure[Long-horizon Tasks]{\label{fig:example-ours}\includegraphics[scale=0.32]{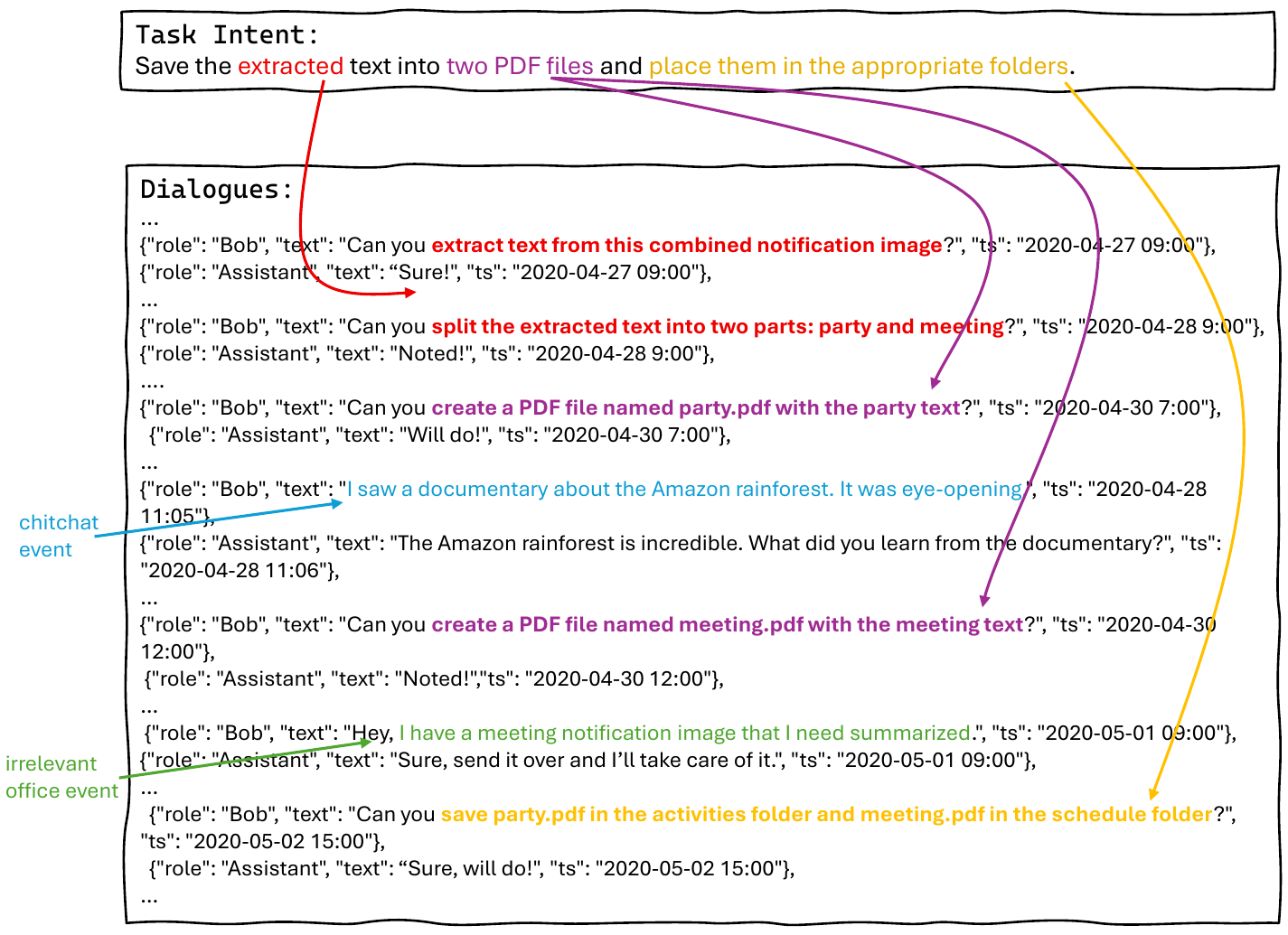}} 
\caption{(a) Atomic tasks: each task is self-contained and does not rely on previous interactions or context. (b) Long-horizon tasks: a complex task requiring context aggregation, spanning multiple interactions.}
\label{fig:visualize-example}
\end{figure}

However, existing benchmarks for agents predominantly focus on atomic tasks that are self-contained and independent of previous interactions or accumulated context \citep{zhou2023webarena,paranjape2023cross,bonatti2024windows,wang2024officebench,xu2024theagentcompany}, as illustrated in \autoref{fig:example-others}. While these benchmarks serve as valuable initial assessments, they fundamentally misrepresent the nature of real-world workflows, which typically unfold across extended periods and encompass various agent-user interactions and require agents to systematically curate, integrate, and leverage information accumulated over extended periods \citep{schick2023toolformer,wang2024jarvis,hu2024hiagent,erdogan2025plan}. Agents that perform well on atomic task benchmarks may struggle with the contextual dependencies, information persistence, and collaborative workflow management required in real-world scenarios.

In this work, we address these challenges by introducing a novel benchmark \textbf{\datasetall} designed to evaluate agents on complex, long-horizon workflows spanning diverse office applications, including \includegraphics[height=1em]{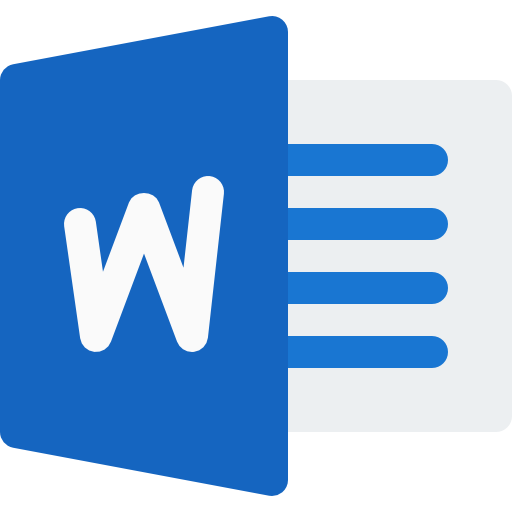} Word, \includegraphics[height=1em]{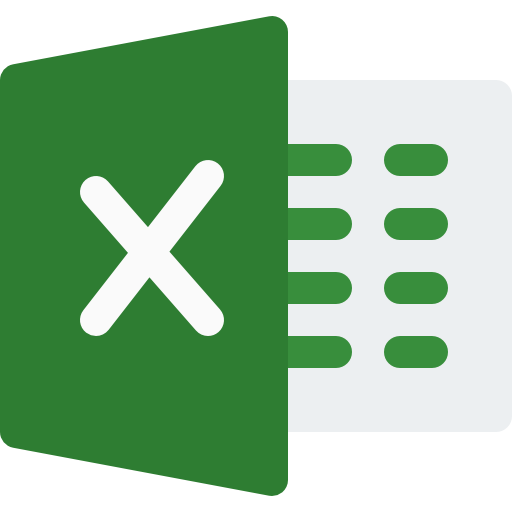} Excel, \includegraphics[height=1em]{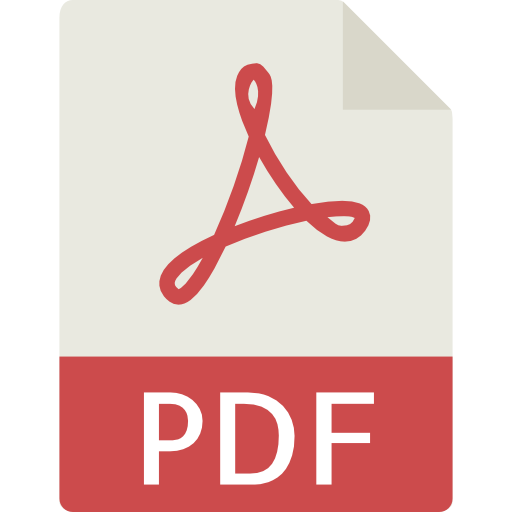} PDF, \includegraphics[height=1em]{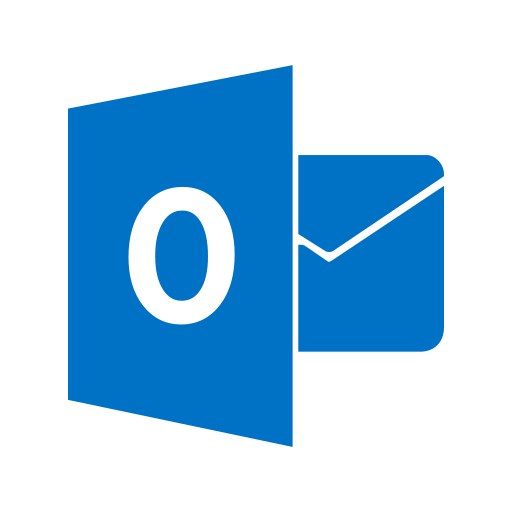} Email, and \includegraphics[height=1em]{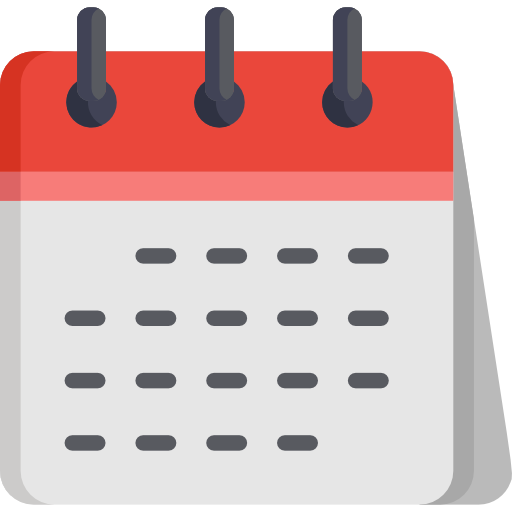} Calendar. Our benchmark includes two splits: \textbf{\datasetold}, which consists of 300 long-horizon tasks originated from real-world use cases in OfficeBench \citep{wang2024officebench}, and \textbf{\datasetnew}, which contains 302 newly generated tasks that are more complex and diverse. 
Each task, as illustrated in \autoref{fig:example-ours}, is designed to require the agent to reason about the task and extract essential information from long-horizon dialogue histories between the user and agent. This enables the construction of feasible workflows and supports multi-step reasoning across various applications. The tasks are structured to reflect the complexities of agent-user interactions, emphasizing the need for agents to maintain context, synthesize information from prior exchanges, and coordinate actions across diverse tools and environments.

Furthermore, many benchmarks rely on costly human annotation, limiting scalability and constraining the diversity of evaluation scenarios \citep{zhou2023webarena,xu2024theagentcompany,yao2024tau}. While recent efforts have explored synthetic data generation with LLMs \citep{ou2024synatra,xu2024agenttrek,xie2025agentsynth}, these approaches typically yield atomic tasks, lacking the sustained interactions and long-term context essential for realistic workflows. These limitations highlight the urgent need for systematic, automated benchmarks that accurately reflect the challenges of real-world, long-horizon tasks.

To address these challenges, we propose \textbf{\ours}, a multi-agent framework that automates the generation of long-horizon workflow benchmarks. Our framework consists of two complementary components: \textbf{\oursold}, which leverages existing benchmarks from OfficeBench \citep{wang2024officebench} and employs a two-agent iterative refinement process to transform atomic tasks into contextually rich, multi-interaction scenarios, thereby creating \datasetold; and \textbf{\oursnew}, which utilizes a multi-agent system operating within realistic application environments to generate entirely new long-horizon tasks from scratch, producing \datasetnew. Through systematic environment exploration, task generation, and dialogue creation, \ours enables scalable production of diverse, contextually grounded benchmark tasks that reflect the complexity of real-world productivity scenarios while maintaining the quality standards necessary for rigorous agent evaluation.

We conduct extensive evaluations of \datasetall using state-of-the-art agents, demonstrating that these benchmarks effectively challenge current models and provide a more accurate assessment of their capabilities in real-world contexts.

In summary, our contributions are as follows:
\begin{itemize}
    \item We introduce \textbf{\datasetall}, a comprehensive benchmark for evaluating agents on long-horizon workflows across multiple office applications, consisting of \textbf{\datasetold} and \textbf{\datasetnew}.
    \item We propose \textbf{\ours}, a multi-agent framework that automates the generation of long-horizon tasks, enabling scalable and diverse benchmark creation.
    \item We demonstrate the effectiveness of \datasetall in challenging state-of-the-art language agents, providing insights into their performance in complex, real-world scenarios.
    \item We analyze the impact of dialogue storage formats within \datasetall, demonstrating that semantic compression and coherent aggregation are essential for effective multi-step reasoning and agent performance.
\end{itemize}

\section{Related Work}

\paragraph{Evaluating LLMs in Executive Environments}

As LLMs advance in tackling real-world challenges \citep{hurst2024gpt,jaech2024openai,openai2025o3-o4mini,anthropic2025claude4,anthropic2025claude37sonnet,comanici2025gemini}, there is a growing shift toward evaluating their capabilities in dynamic, executive environments rather than static datasets. Beyond text-based games \citep{cote2018textworld,shridhar2020alfworld}, recent research increasingly simulates realistic scenarios to assess agents' proficiency in tool use \citep{deng2023mind2web,qin2023webcpm,zhuang2023toolqa,qin2023toolllm,lu2024weblinx,wang2024executable,shen2024shortcutsbench,xu2024theagentcompany,sutela2024game}. Current benchmarks, such as WebArena \citep{zhou2023webarena}, AgentBench \citep{paranjape2023cross},  WindowsArena \citep{bonatti2024windows}, and OfficeBench \citep{wang2024officebench}, provide valuable evaluation settings focused on web and office environments. However, these platforms primarily measure atomic performance in self-contained contexts and lack mechanisms to evaluate LLM agents' interactions with complex environments over extended periods. This limitation is significant, as robust assessment of planning, long-term information retrieval, and execution is essential for understanding agents' true capabilities in real-world tasks.

\paragraph{Synthetic Benchmark Generation}

Existing agent datasets and benchmarks largely rely on human annotators for task creation, demonstrations, and evaluation metric design \citep{zhou2023webarena,xu2024theagentcompany,yao2024tau}, resulting in high costs and limited diversity. Recent studies try to leverage LLMs to automatically generate agent tasks and trajectories \citep{ou2024synatra,xu2024agenttrek,xie2025agentsynth}. For instance, \citet{murty2024nnetnav,pahuja2025explorer,trabucco2025insta,gandhi2025go} employ LLMs as web agents to synthesize web-based interactions in semi-realistic environments. Moreover, composing atomic tasks is another method to construct more challenging tasks \citep{boisvert2024workarena++,drouin2024workarena}. \citet{li2024autobencher} iteratively propose and refine dataset descriptions to generate topic-specific problems. However, these approaches predominantly focus on web-based activities and are generally limited to simple interactions, lacking the complexity of multi-step reasoning and extensive tool use required for robust agent evaluation.

\paragraph{Ours}

Distinct from previous approaches, we introduces a multi-agent framework \ours to automatically construct the long-term workflow benchmark \datasetall, enabling a more rigorous assessment of agents' abilities to curate context to handle complex tasks. \datasetall is specifically designed to evaluate agent performance in realistic office scenarios, where agents must interact with multiple applications to accomplish intricate objectives. This benchmark challenges agents to reason about \intent, extract critical information from dialogue history, and assemble feasible workflows, thereby providing a comprehensive evaluation of their capabilities in dynamic, multi-step environments.
\begin{figure*}
    \centering
    \includegraphics[width=\linewidth]{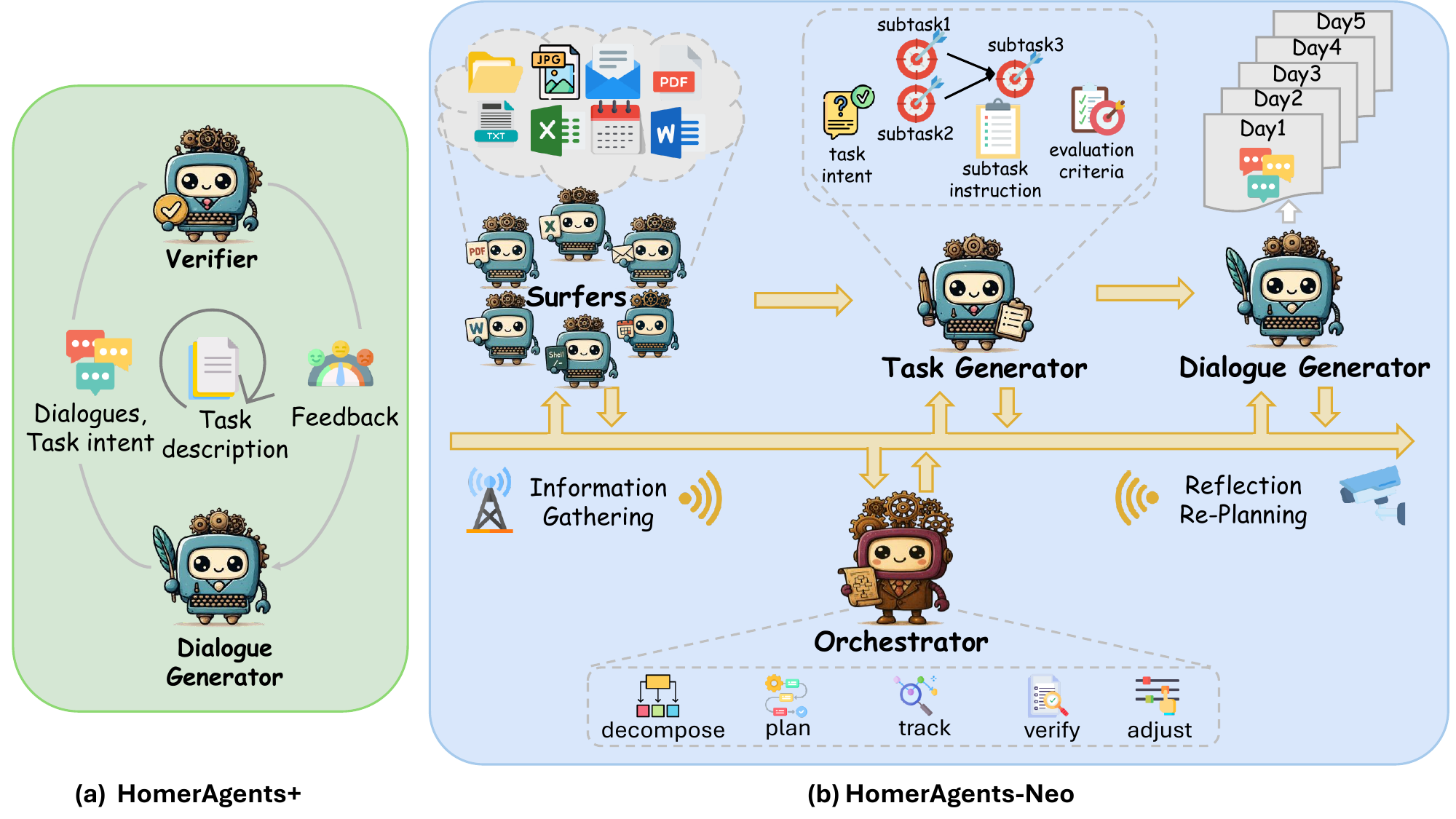}
    \caption{\ours Framework Overview. \ours consists of two components: \oursold and \oursnew. \oursold builds upon the task descriptions from OfficeBench to generate long-horizon dialogues, while \oursnew creates entirely new tasks and corresponding dialogues from scratch by employing a multi-agent system that operates within realistic application environments.}
    \label{fig:magentic-framework}
\end{figure*}

\begin{algorithm}[t]
\small
\caption{\oursold}
\label{alg:oursold}
\KwIn{Task description $\mathcal{T}$; the generator $\mathcal{G}$; the verifier $\mathcal{V}$; the maximal number of iterations $N_{\text{max}}$;}
\KwOut{Task intent $\mathbb{I}$ and dialogues $\mathbb{D}$;}
$\mathbb{F}_{0} \leftarrow \varnothing$ \Comment*[r]{Initialize empty feedback}
\For{i=1 to $N_{\text{max}}$}{
    $\{\mathbb{I}_i, \mathbb{D}_i\} \leftarrow \mathcal{G}(\mathcal{T}, \mathbb{F}_{i-1})$ \Comment*[r]{The generator $\mathcal{G}$ generates the task intent $\mathbb{I}$ and dialogues $\mathbb{D}$}
    $\mathbb{F}_i \leftarrow \mathcal{V}( \mathcal{T}, \mathbb{I}_i, \mathbb{D}_i)$ \Comment*[r]{The verifier $\mathcal{V}$ evaluates $\mathbb{I}$ and $\mathbb{D}$, and provides feedback $\mathbb{F}_i$ }
    \If{$\mathbb{F}_i$ == pass}{
        \Return{ $\{\mathbb{I}_i, \mathbb{D}_i\} $} \Comment*[r]{Early stop if the verifier $\mathcal{V}$ thinks the task intent $\mathbb{I}$ and dialogues $\mathbb{D}$ are satisfactory}
    }
}
\Return{$\{\mathbb{I}_{N_{\text{max}}}, \mathbb{D}_{N_{\text{max}}}\}$}  \Comment*[r]{Return the task intent $\mathbb{I}$ and dialogues $\mathbb{D}$ after $N_{\text{max}}$ iterations}
\end{algorithm}

\begin{algorithm}[t]
\small
\caption{\oursnew}
\label{alg:oursnew}
\KwIn{Applications $\mathcal{A}=\{a_k\}_{k=0}^{K}$; Environment $\mathcal{E}$;  Orchestrator $\mathcal{O}$; Surfers $\mathcal{S}=\{S_k\}_{k=0}^{K}$; Task Generator $\mathcal{G}_{\text{task}}$; Dialogue Generator $\mathcal{G}_{\text{dial}}$;}
\KwOut{Task $\tau$ and dialogue $\mathbb{D}$;}

\textbf{Phase 1: Planning};

$\mathbb{P} \leftarrow \mathcal{O}(\mathcal{A}, \mathcal{E})$ where $\mathbb{P}=\{ \mathbb{P}_{\text{surf}}, \mathbb{P}_{\text{task}}, \mathbb{P}_{\text{dial}} \}$\Comment*[r]{Orchestrator drafts the generation plan $\mathbb{P}$}

\textbf{Phase 2: Environment Exploration}\;

$\mathbb{C} \leftarrow \bigcup_{k=0}^{K} S_k(\mathbb{P}_{\text{surf}}, a_k, \mathcal{E})$\Comment*[r]{Surfers collect contextual information from environment $\mathcal{E}$}

\textbf{Phase 3: Task Generation};

$\tau \leftarrow \mathcal{G}_{\text{task}}(\mathbb{P}_{\text{task}}, \mathbb{C})$ where $\tau = \{\mathbb{T}, \mathbb{I}, \mathbb{K}, \mathbb{E}\}$ \Comment*[r]{Task Generator generate task components, including task description $\mathbb{T}$, task intent $\mathbb{I}$, subtask instructions $\mathbb{K}$, and evaluation criteria $\mathbb{E}$}

\textbf{Phase 4: Dialogue Generation};

$\mathbb{D} \leftarrow \mathcal{G}_{\text{dial}}(\mathbb{P}_{\text{dial}}, \mathbb{C}, \mathbb{I},\mathbb{K})$   \Comment*[r]{{Dialogue generator generates T-Days  dialogues}}

\Return{Task $\tau$ and dialogues $\mathbb{D}$ }\Comment*[r]{Complete task for dataset}
\end{algorithm}

\section{Methodology}
\label{sec:method}

In this section, we firstly introduce \ours, a multi-agent framework that automatically generates the long-horizon workflow benchmark \datasetall in \autoref{sec:method_method}, including two components: \oursold (\autoref{sec:method_old}) and \oursnew (\autoref{sec:method_new}). We then describe the long-horizon workflow benchmark \datasetall in \autoref{sec:method_benchmark}, including the dataset analysis (\autoref{sec:method_analysis}), quality control measures (\autoref{sec:method_quality_control}), and human evaluation (\autoref{sec:method_human_evaluation}).

\subsection{\ours: Automating Benchmark Creation}
\label{sec:method_method}

It is highly challenging to create \datasetall in a scalable and reliable manner, as it requires generating realistic user–assistant interaction histories and the context-dependent multi-step tasks that reflect the complexity and ambiguity of real-world productivity scenarios. To facilitate this process, we propose a multi-agent framework \ours that automates the generation of \datasetall benchmark tasks, including \oursold (see \autoref{sec:method_old}) and \oursnew (see \autoref{sec:method_new}). 

\subsubsection{\oursold: Standing on the Shoulders of \officebench}
\label{sec:method_old}

\oursold builds upon the task descriptions from \officebench \citep{wang2024officebench} to generate long-horizon dialogue scenarios that more closely mirror real-world productivity workflows. Starting from a given task description $\mathcal{T}$, \oursold employs a two-agent iterative refinement framework to produce task intents $\mathbb{I}$ and corresponding long-horizon user-assistant dialogues $\mathbb{D}$, thereby contextualizing and enriching the original task.

The framework comprises two core components: a \textbf{generator} ($\mathcal{G}$) and a \textbf{verifier} ($\mathcal{V}$), as depicted in \autoref{fig:magentic-framework}. The generator $\mathcal{G}$ receives the task description $\mathcal{T}$ and any feedback from previous iterations $\mathbb{F}_{i-1}$, and outputs a task intent $\mathbb{I}_i$ along with a corresponding dialogue $\mathbb{D}_i$. Here, the task intent $\mathbb{I}$ succinctly captures the user's goal without specific details, while the dialogue $\mathbb{D}$ provides the natural conversational context leading to the task. The verifier $\mathcal{V}$ then evaluates the generated content against criteria such as dialogue realism, task alignment, and contextual coherence, returning structured feedback $\mathbb{F}_i$.

This process is executed iteratively, as outlined in \autoref{alg:oursold}, with a maximum of $N_{\text{max}}$ iterations. In each iteration $i$, the generator $\mathcal{G}$ refines its output based on the original task and accumulated feedback, while the verifier $\mathcal{V}$ either approves the result (``pass'') or provides actionable feedback for further improvement. The cycle continues until the verifier approves the content or the iteration limit is reached. By leveraging established benchmarks and introducing an iterative, feedback-driven process, \oursold enables the creation of contextually grounded, long-horizon tasks that are both practically relevant and sufficiently complex to rigorously evaluate long-horizon workflow task understanding in productivity settings.

\subsubsection{\oursnew: Scaling up the Benchmark Creation}
\label{sec:method_new}

While \oursold effectively leverages existing benchmarks, \oursnew addresses the need for more diverse and scalable task generation by creating entirely new long-horizon tasks from scratch. \oursnew employs a multi-agent system that operates within realistic application environments to generate authentic productivity scenarios, as shown in \autoref{fig:magentic-framework}.

\oursnew consists of \textbf{productivity applications} $\mathcal{A}=\{a_k\}_{k=0}^{K}$, \textbf{environment} $\mathcal{E}$, \textbf{orchestrator} $\mathcal{O}$, \textbf{surfers} $\mathcal{S}=\{S_k\}_{k=0}^{K}$, \textbf{task generator} $\mathcal{G}_{\text{task}}$, and \textbf{dialogue generator} $\mathcal{G}_{\text{dial}}$. Orchestrator $\mathcal{O}$ manages planning, progress tracking, and coordinates the entire generation process by orchestrating each stage of data generation, ensuring coherence in both task and dialogue creation. Surfers $\mathcal{S}$ gather information from environment by interacting with a diverse set of simulated productivity applications. Task generator $\mathcal{G}_{\text{task}}$ synthesizes the tasks and corresponding evaluation criteria. Dialogue generator $\mathcal{G}_{\text{dial}}$ then creates multi-day dialogues simulating realistic user-assistant interactions.

The framework consists of four distinct phases, as outlined in \autoref{alg:oursnew}:

\paragraph{Phase 1: Planning} The orchestrator $\mathcal{O}$ receives a set of applications $\mathcal{A} = \{a_k\}_{k=0}^{K}$ and environment $\mathcal{E}$, then formulates a comprehensive generation plan $\mathbb{P} = \{\mathbb{P}_{\text{surf}}, \mathbb{P}_{\text{task}}, \mathbb{P}_{\text{dial}}\}$. This plan specifies how the subsequent phases should explore the environment $\mathbb{P}_{\text{surf}}$, generate tasks $\mathbb{P}_{\text{task}}$, and create dialogues $\mathbb{P}_{\text{dial}}$.

\paragraph{Phase 2: Environment Exploration} A collection of specialized surfers $\mathcal{S} = \{S_k\}_{k=0}^{K}$ systematically explore the application environment. Each surfer $S_k$ follows the surfing plan $\mathbb{P}_{\text{surf}}$ to interact with application $a_k$ within environment $\mathcal{E}$, collecting contextual information $\mathbb{C}$. This exploration phase ensures that generated tasks are grounded in realistic application capabilities and user workflows.

\paragraph{Phase 3: Task Generation} The task generator $\mathcal{G}_{\text{task}}$ utilizes the collected contextual information $\mathbb{C}$ and the task generation plan $\mathbb{P}_{\text{task}}$ to create comprehensive task specifications $\tau = \{\mathbb{T}, \mathbb{I}, \mathbb{K}, \mathbb{E}\}$. This includes the task description $\mathbb{T}$, the task intent $\mathbb{I}$, detailed subtask instructions $\mathbb{K}$, and evaluation criteria $\mathbb{E}$. The task description $\mathbb{T}$ outlines the specific goals and requirements of the task, the task intent $\mathbb{I}$ conveys the high-level overall goal but omits specific details of the task, $\mathbb{K} = \{k_1, \ldots, k_t\}$ provides instructions for completing the task, and the evaluation criteria $\mathbb{E}$ define how the task's success will be measured.

\paragraph{Phase 4: Dialogue Generation} Finally, the dialogue generator $\mathcal{G}_{\text{dial}}$ creates natural user-assistant conversations $\mathbb{D}$ that lead to the generated task. This process incorporates the dialogue plan $\mathbb{P}_{\text{dial}}$, contextual information $\mathbb{C}$, task intent $\mathbb{I}$, and subtask instructions $\mathbb{K}$ to produce realistic long-horizon dialogues that capture the gradual evolution of user requirements. For each subtask instruction $k_i \in \mathbb{K}$, the dialogue generator $\mathcal{G}_{\text{dial}}$ produces a corresponding dialogue $\mathbb{D}_i$ that simulates the interaction between the user and the assistant, reflecting how the task is approached over multiple days. Combining these dialogues, we obtain a comprehensive dialogue history $\mathbb{D} = \{\mathbb{D}_1, \ldots, \mathbb{D}_t\}$ that illustrates the user's journey through the task. Additionally, we also include task-irrelevant content (e.g. chitchat) in the generated dialogues $\mathbb{D}$ to make the generated content align better with the real-world scenarios.

By decomposing the generation process into these four phases, \oursnew ensures systematic exploration of application environments while maintaining coherence between the generated tasks and dialogues. This approach enables scalable creation of diverse, contextually grounded benchmark tasks that reflect the complexity of real-world productivity scenarios.

\subsubsection{Implementation Details}
\label{sec:method_implementation}

Considering the trade-off between performance and cost, we implement all agents in \ours using the GPT-4.1 model, which offers strong capabilities for complex reasoning tasks while remaining cost-effective. The maximum iterations $N_{\text{max}}$ in \autoref{alg:oursold} are set to 5, allowing for sufficient exploration of the task space while managing computational resources effectively. The $T$ in \autoref{alg:oursnew} is set to 5, representing generating at least five days of dialogues, which is sufficient to capture the complexity of long-term workflows. Additionally, we implement \oursnew based on the Magentic-One framework~\citep{fourney2024magentic}.

During dialogue generation, when the user assigns subtasks to the assistant, the assistant does not actually execute the tasks but instead simulates execution by generating responses based on the task descriptions and dialogue context. This approach enables us to focus on generating diverse and realistic dialogues at scale, without the need for real task execution. By deferring execution, our benchmark evaluates agents' abilities to curate and integrate information distributed across multiple dialogue turns and days, an essential aspect for assessing long-horizon comprehension and planning.

\subsection{\datasetall: Long-Horizon Workflow Benchmark}
\label{sec:method_benchmark}

\subsubsection{Evaluation}
\label{sec:data_evaluation}

In \datasetall, LLM agents are required to interact with multiple applications to complete complex tasks. This process demands that agents reason about task intent and extract essential information from the dialogue history to construct feasible workflows. We construct \datasetall within a Docker environment containing pre-installed applications and automate operations using Python libraries. We set up a file system to manage documents, emails, and calendar events required for the tasks. After the agents complete each task, we save the entire file system and perform customized evaluations to verify correctness.

Our evaluation integrates exact matching, fuzzy matching, and execution-based methods. Exact and fuzzy matching assess whether the agent's task output aligns with the expected (e.g., keyword matching for generated documents and calendar events), while the execution-based method verifies if the agent's task outputs can be successfully evaluated via code snippets (e.g., checking calendar conflicts). The output is considered to be successful if all the evaluation criteria are satisfied.  We report the \textbf{pass rate} as the measure of model performance, where the pass rate is defined as the percentage of tasks completed successfully: $\frac{\#\text{successful tasks}}{\#\text{total tasks}}$.


\begin{table}[t] 
\centering
\small
\setlength{\tabcolsep}{2pt}
\caption{
Data statistics of \datasetold and \datasetnew.}
\begin{tabular}{lcccccccc}
\toprule
& \multicolumn{4}{c}{\datasetold}   & \multicolumn{4}{c}{\datasetnew}   \\ \cmidrule(rl){2-5} \cmidrule(rl){6-9}
& \begin{tabular}[c]{@{}c@{}}single\\ apps\end{tabular} & \begin{tabular}[c]{@{}c@{}}two\\ apps\end{tabular} & \begin{tabular}[c]{@{}c@{}}three\\ apps\end{tabular} & \textbf{overall} & \begin{tabular}[c]{@{}c@{}}single\\ apps\end{tabular} & \begin{tabular}[c]{@{}c@{}}two\\ apps\end{tabular} & \begin{tabular}[c]{@{}c@{}}three\\ apps\end{tabular} & \textbf{overall} \\ \midrule
Total \# conversation $h.$       & \phantom{000}93        & \phantom{000}95     & \phantom{00}112      & \phantom{00}300  & \phantom{000}60        & \phantom{000}71     & \phantom{00}171      & \phantom{000}302  \\
Avg. \# session $k.$ in conversation $h$    & \phantom{00}27.8        & \phantom{00}24.7     & \phantom{00}30.6       & \phantom{00}27.9   & \phantom{000}5.0         & \phantom{000}5.0      & \phantom{000}5.1        & \phantom{000}5.0    \\
Avg. \# utterance $j.$ in session $k$    & \phantom{00}10.8        & \phantom{00}12.1     & \phantom{00}11.4       & \phantom{00}11.4   & \phantom{00}72.3        & \phantom{00}73.5     & \phantom{00}73.3       & \phantom{00}73.2  \\ \midrule
Avg. \# tokens. conversation $h$ & 3323.2      & 3209.6   & 3809.9     & 3468.9 & 5031.6      & 5223.1   & 5196.4     & 5169.9 \\
Avg. \# tokens. sessions $k$    & \phantom{0}119.7       & \phantom{0}130.1    & \phantom{0}124.4      & \phantom{0}124.6  & 1006.3      & 1041.7   & 1026.1     & 1025.8 \\
Avg. \# tokens. utterance $j$   & \phantom{00}11.1        & \phantom{00}10.8     & \phantom{00}10.9       & \phantom{00}10.9   & \phantom{00}13.9        & \phantom{00}14.2     & \phantom{00}14.0       & \phantom{00}14.0   \\
\bottomrule
\end{tabular}
\label{tab:statistics}
\end{table}

\begin{figure}[t]
\centering          
\subfigure[Execution steps]{\label{fig:execution-turns}\includegraphics[scale=0.6]{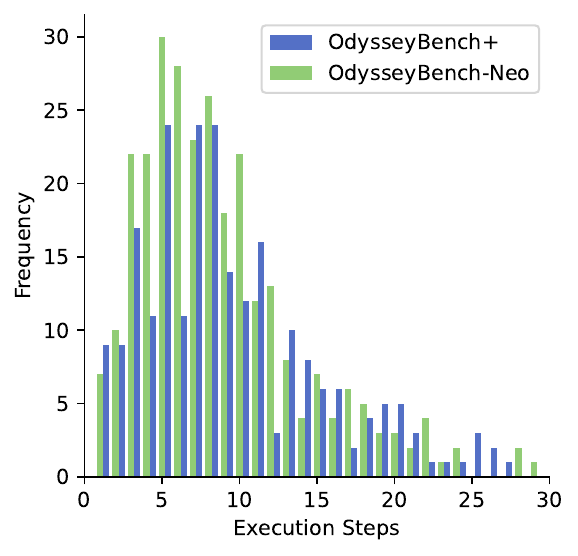}} 
\subfigure[Verb-Noun-Apps]{\label{fig:verb-noun}\includegraphics[scale=0.23]{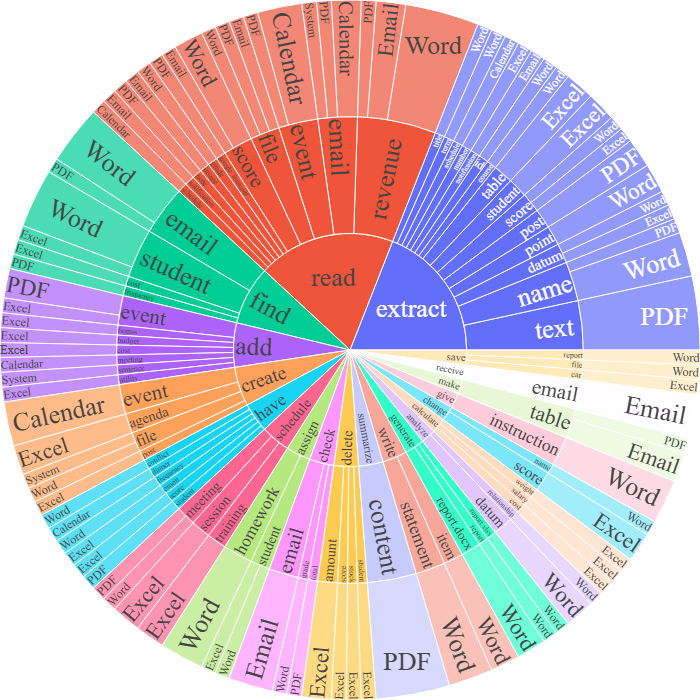}} 
\caption{
    (1)  Execution steps needed for the tasks in \datasetall. (b) Actions, objects, and applications of \datasetall.
}
\label{fig:visualize-statisticsexample}
\end{figure}

\subsubsection{Dataset Analysis}
\label{sec:method_analysis}

As shown in \autoref{tab:statistics}, our dataset comprises 602 tasks, categorized by the number of applications involved: Single App (153 tasks), Two Apps (166 tasks), and Three Apps (283 tasks). Each task is documented through multi-day dialogues, with at least five days per task. Dialogues occurring within the same day are grouped into a single session, and every dialogue contains a minimum of 10 utterances, ensuring rich interaction data. \datasetold contains 300 conversation histories with an average of 27.9 sessions per conversation and 11.4 utterances per session, resulting in relatively short sessions with an average of 124.6 tokens per session. In contrast, \datasetnew comprises 302 conversations with a more structured format of exactly 5 sessions per conversation (corresponding to the 5-day dialogue design) but significantly longer sessions, averaging 1025.8 tokens each and 73.2 utterances per session. This design difference reflects \datasetnew's focus on creating more comprehensive daily interactions, while \datasetold maintains the original fragmented conversation structure from \officebench. Overall, \datasetnew generates richer conversational content with approximately 49\% more tokens per conversation (5169.9 vs. 3468.9 tokens), demonstrating the enhanced depth and complexity of the newly generated tasks.

We further analyze the distribution of execution steps in \datasetold and \datasetnew, as illustrated in \autoref{fig:execution-turns}. The number of execution steps required to complete a task is consistent across both datasets, with the majority of tasks requiring 3-15 execution turns. This demonstrates that tasks in \datasetall are sufficiently complex, mirroring real-world scenarios where users must navigate multi-step workflows across multiple applications. Furthermore, we also analyze the diversity of the tasks in \datasetall, an overview of actions, objects, and applications in \datasetall is provided in \autoref{fig:verb-noun}. Our \datasetall benchmark encompasses a wide range of actions, objects, and applications, ensuring that it captures the complexity and variety of real-world productivity tasks. This diversity enhances the benchmark's applicability to various productivity scenarios, making it a valuable resource for evaluating long-horizon workflow understanding in LLMs.

\subsubsection{Quality Control}
\label{sec:method_quality_control}

\begin{table}[t]
\centering
\small
\setlength{\tabcolsep}{2pt}
\caption{Quality verification performance of generated \intent and dialogues for \datasetall.}
\begin{tabular}{lcc}
\toprule
Metric            & \datasetold & \datasetnew \\ \midrule
Completeness & 81.33  & 93.71   \\
Appropriateness  & 88.33   & 88.08  \\
Both   & 72.67  & 83.77 \\ \bottomrule
\end{tabular}
\label{tab:quality}
\end{table}

\paragraph{Automated Validation} 
 
To ensure the generated tasks are both high-quality and solvable, we implement a systematic automated validation pipeline. Our approach consists of two primary validation stages. First, we verify task solvability by filtering out tasks whose evaluation criteria $\mathbb{E}$ fall outside our predefined evaluation function library. This ensures that each task has well-defined, measurable success criteria. Second, we conduct a consistency check between the task description $\mathbb{T}$ and the information available to agents during evaluation. Specifically, we test whether a powerful LLM (o3) can solve the task when provided with: (1) the task description $\mathbb{T}$, and (2) only the task intent $\mathbb{I}$ and subtask instructions $\mathbb{K}$. Tasks are considered valid only if the LLM succeeds in both scenarios, ensuring that the dialogue contains sufficient information for task completion while the intent appropriately abstracts the core objective. This cross-validation eliminates under-specified tasks and confirms that essential information is properly embedded within the conversational context. To complement this automated filtering, we employ an LLM-as-a-judge method where five independent GPT-4.1 agents evaluate the generated task intent $\mathbb{I}$ and dialogues $\mathbb{D}$ across two key dimensions:

\begin{enumerate}
    \item \textbf{Completeness}: For any given task description, the combination of \intent and dialogues should provide sufficient information for an agent to solve the task without omitting any necessary details. 
    \item \textbf{Soundness}: The \intent should not leak any specific information from the task description; all essential details must be conveyed through the dialogues. 
\end{enumerate}

Each agent provides an independent judgment, and a majority voting mechanism aggregates these assessments to determine the overall data quality. The results of this evaluation are presented in \autoref{tab:quality}. A minimum pass rate of 70\% reflects the quality of the data, indicating the effectiveness of the automatic generation of \ours.

\paragraph{Human Verification and Post-Editing}
In addition to automated validation, we implement human curation to further enhance the quality of the generated \intent and dialogues. A team of three native English-speaking annotators manually reviews the generated \intent and dialogues, assessing them for completeness, appropriateness, and logical coherence. During this process, curators remove any tasks that fail to meet established quality standards. While our cross-checking mechanism filters out tasks deemed unsolvable by the LLM, this method is inherently limited by the LLM's problem-solving capabilities. Consequently, human curation intentionally includes tasks that satisfy quality standards yet remain unsolvable by the LLM. This human-in-the-loop approach ensures that the resulting dataset is both challenging and reflective of real-world use cases.

\subsubsection{Human Evaluation}
\label{sec:method_human_evaluation}

\begin{table} \small
    \centering
    \begin{tabular}{lcccc}
    \toprule
        Task & 1-apps  & 2-apps & 3-apps & overall \\ \midrule
        Human & 92.31 & 90.00 & 91.67 & 91.43 \\ \bottomrule
    \end{tabular}
    \caption{Human performance of \oursnew}
    \label{tab:human}
\end{table}
We also ask two human annotators to perform a randomly sampled subset of the tasks and report the human performance in \autoref{tab:human}. The human annotators are asked to complete the tasks using the same productivity applications as those provided to the agents. The human performance is over 90\%, indicating that the tasks are solvable and coherent.

\section{Experimental Setup}
\label{sec:data_experiment_setup}

\paragraph{Long-Context Evaluation}
We evaluate the agent performance on \datasetall using the long-context setting, where the entire dialogue history is provided to the agent. 

\paragraph{RAG Evaluation}
We also evaluate agent performance on \datasetall under the Retrieval-Augmented Generation (RAG) setting, where the agent retrieves relevant context from the dialogue history using embedding models to generate responses. We conduct experiments with two types of stored context: (1) \textbf{raw context} and (2) \textbf{summarized context}. Furthermore, each type of context is organized into two levels of granularity. For raw context: (a) \emph{session-level:} the entire dialogue of each session is stored and embedded as a single document; (b) \emph{utterance-level:} each user/assistant turn is treated as a separate document and embedded independently. For summarized context: (a) \emph{session-level:} the full session is summarized and stored as a single document; (b) \emph{chunk-level:} multiple sessions are concatenated and segmented into coherent chunks, with each chunk summarized independently.

\paragraph{Evaluation Metrics}
As mentioned in \autoref{sec:data_evaluation}, we measure the agent performance using the \textbf{pass rate}, which is the percentage of successful task completions out of the total number of tasks.

\paragraph{Models}
We evaluate the long-horizon workflow automation capabilities of the agents of the proprietary LLMs, including o3, o3-mini, GPT-4o, GPT-4o-mini, GPT-4.1, GPT-5, and GPT-5-chat, and the open-weight LLMs, including DeepSeek-R1, DeepSeek-R1-Distill-Qwen-32b, and Qwen3-32b, as these models are among the highest-ranking LLMs available. And the embedding model used for RAG is OpenAI text-embedding-3-large.

\begin{table}[t] 
\centering
\small
\setlength{\tabcolsep}{5pt}
\caption{
Main results given by multiple proprietary models or open-weight models on \datasetold and \datasetnew tasks under the long-context configuration. We divide the tasks into “1/2/3-apps”, specifying the number of applications required by the tasks. The overall performance is reported as the macro-average across all tasks.}
\begin{tabular}{lcccccccc}
\toprule
& \multicolumn{4}{c}{\datasetold}   & \multicolumn{4}{c}{\datasetnew}   \\ \cmidrule(rl){2-5} \cmidrule(rl){6-9}
& 1-apps & 2-apps & 3-apps & \textbf{overall} & 1-apps & 2-apps & 3-apps & \textbf{overall} \\ \midrule
\multicolumn{9}{l}{\cellcolor{gray!15}\textbf{Proprietary Models}}  \\ 
o3   & 72.83   & \textbf{70.53}& \textbf{30.36}  & \textbf{56.19}& 68.33   & 60.56& \textbf{59.06}  & \textbf{61.26}\\
o3-mini  & 38.04   & 20.00& 15.18  & 23.75& 71.67   & 39.44& 45.61  & 49.34\\
GPT-4o-mini  & 30.11   & 22.11& \phantom{0}7.14   & 19.00& 65.00   & 33.80& 29.83  & 37.75\\
GPT-4o   & 47.31   & 42.11& 15.18  & 33.67& \textbf{75.00}   & 47.89& 45.61  & 51.99\\
GPT-4.1  & 55.91   & 43.16& 12.50  & 35.67& \textbf{75.00}   & \textbf{63.38}& 47.37  & 56.62\\
GPT-5-chat & 55.91 & 48.42 &  20.54 & 40.33 & \textbf{75.00} & 57.75 & 51.46 & 57.62\\
GPT-5 & \textbf{75.27} & 66.32 & 25.89 & 54.00 &  61.67 & 56.34  &  53.80 & 55.96    \\ 
\multicolumn{9}{l}{\cellcolor{gray!15}\textbf{Open-weight Models}}  \\ 
DeepSeek-R1  & \textbf{53.76}   & \textbf{47.37}& \textbf{20.54}  & \textbf{39.33}& \textbf{78.33}   & \textbf{60.56}& \textbf{44.44}  & \textbf{54.97}\\
DS.-Distill-Qwen-32b & 30.11   & 16.84& \phantom{0}1.79   & 15.33& 40.00   & 22.54& 10.53  & 19.21\\
Qwen-3-32b & 38.71   & 33.68& 11.61  & 27.00   & 41.67 & 22.54 & 21.05 & 25.50 \\
\bottomrule
\end{tabular}
\label{tab:long-context}
\end{table}

\begin{table}[t] 
\centering
\small
\setlength{\tabcolsep}{5pt}
\caption{
Performance of RAG-based GPT-4o on the \datasetold. ``Long-context prompting baseline'' represents the results evaluated in the long-context setting. ``top-k'' means the top-k retrieved documents used as the context, and ``tokens'' indicates the number of tokens in the retrieved documents. }
\begin{tabular}{cccccccc}
\toprule
\multicolumn{1}{l}{Storage} & granularity      & top-k & tokens & 1-apps & 2-apps & 3-apps & overall \\ \midrule
 \multicolumn{3}{l}{Long-context prompting baseline}      & 8000     & 47.31       & \textbf{42.11}    & 15.18      & \textbf{33.67 }  \\ \midrule
\multirow{6}{*}{raw} & \multirow{2}{*}{session}   & 5  & 750    & 40.86       & 40.00    & 11.61      & 29.67   \\
&& 10 & 1500   & 39.79       & 40.00    & 14.29      & 30.33   \\  \cdashline{2-8}
& \multirow{4}{*}{utterance} & 5  &    80    & 29.03       & 35.79    & \phantom{0}8.04       & 23.33   \\
& & 10 &   155     & 27.96       & 33.68    & \phantom{0}8.93       & 22.67   \\
&& 25 &    370    & 39.79       & 35.79    & 12.50      & 28.33   \\
&& 50 &    730    & \textbf{57.69}       & 40.00    & 17.17      & 29.41   \\ \midrule
\multirow{6}{*}{summary}    & \multirow{2}{*}{session}   & 5  &  290   & 29.03       & 35.79    & \phantom{0}9.82       & 24.00   \\
&& 10 &  650   & 33.33       & 36.84    & \phantom{0}9.82       & 25.67   \\ \cdashline{2-8}
& \multirow{4}{*}{chunk}    & 5     &   290      & 30.11       & 29.47    & 12.50      & 23.33   \\
    &    & 10    &   380    & 40.86       & 34.74    & 16.96      & 30.00   \\
    &    & 25    &   600      & 46.24       & 36.84    & \textbf{19.64}      & 33.33   \\
&& 50 &   670  &  44.09 & 40.00 & 16.96 & 32.67
   \\
\bottomrule
\end{tabular}
\label{tab:rag-old}
\end{table}

\begin{table}[t] 
\centering
\small
\setlength{\tabcolsep}{5pt}
\caption{Performance of RAG-based GPT-4o on the \datasetnew. ``Long-context prompting baseline'' represents the results evaluated in the long-context setting. ``top-k'' means the top-k retrieved documents used as the context, and ``tokens'' indicates the number of tokens in the retrieved documents.}
\begin{tabular}{cccccccc}
\toprule
\multicolumn{1}{l}{Storage} & granularity      & top-k & tokens & 1-apps & 2-apps & 3-apps  & overall \\ \midrule
\multicolumn{3}{l}{Long-context prompting baseline}       &  6700   & \textbf{75.00}       & 47.89    & 45.61      & 51.99    \\ \midrule
\multirow{4}{*}{raw}& \multirow{4}{*}{utterance}  & 5     &  90   & 30.00       & 16.90    & \phantom{0}8.19       & 14.57   \\
        &            & 10    &  180    & 31.67       & 16.90    & 11.11      & 16.56   \\
        &            & 25    &   450   & 35.00       & 32.39    & 21.05      & 26.49   \\
        &            & 50    &   915   & 56.67       & 40.85    & 31.58      & 38.74   \\  \midrule
\multirow{5}{*}{summary} & session      & 5     &  2200   & \textbf{75.00}       & 46.48    & 49.12      & 53.64   \\ \cdashline{2-8}
 & \multirow{4}{*}{chunk}      & 5     &  1200   & 30.11       & 29.47    & 12.50      & 23.33   \\
   &      & 10    &   1260  & 40.86       & 34.74    & 16.96      & 30.00   \\
        &            & 25    &  1360   & 68.33       & \textbf{59.16}    & \textbf{50.88 }     & \textbf{56.29}  \\
        &            & 50  & 1460  &  68.33 & \textbf{59.16} & 48.54 & 54.97
     \\ 
\bottomrule
\end{tabular}
\label{tab:rag-new}
\end{table}
\begin{table}[t] 
\centering
\small
\setlength{\tabcolsep}{5pt}
\caption{
The number of execution steps of the task in \datasetold and \datasetnew under different configurations indicates how many steps are required to successfully execute the task. ``configuration'' represents the experimental setup used for evaluation.}
\begin{tabular}{lcccccccc}
\toprule
  &  configuration   & 1-apps & 2-apps & 3-apps & overall \\ \midrule
\multirow{3}{*}{\datasetold} & long-context & 6.31      & 11.61  & 12.70    & 10.25       \\
 & RAG-utterance & 6.85      & 11.48  & 14.70    & 11.05       \\
 & RAG-chunk   & 7.25      & 8.28   & 14.86    & 10.10       \\ \midrule
\multirow{3}{*}{\datasetnew} & long-context & 7.81      & 9.63   & 11.74    & 10.46\\
 & RAG-utterance & 8.17      & 9.66   & 12.52    & 10.92       \\
 & RAG-chunk   & 7.93      & 9.92   & 12.54    & 10.95      
 \\ \bottomrule
\end{tabular}
\label{tab:turns}
\end{table}
\section{Experimental Results}
\label{sec:resultls}

In this section, we evaluate several LLMs and RAG-based approaches on the \datasetold and \datasetnew tasks, focusing on their performance across different configurations. We analyze how task complexity, context length, and retrieval strategies impact model effectiveness.

\paragraph{Tasks get increasingly complex with more applications involved, leading to a performance drop.}
As shown in \autoref{tab:long-context}, we observe a consistent performance degradation across all models as the number of applications in a task increases. For \datasetold tasks, the average performance drops from single-app scenarios to three-app scenarios across all models: o3 drops from 72.83 to 30.36, GPT-4.1 from 55.91 to 12.50, and DeepSeek-R1 from 53.76 to 20.54. This trend is also evident in \datasetnew tasks, though the degradation is less severe. For instance, o3 maintains relatively stable performance (68.33 to 59.06), while GPT-4o shows a decline from 75.00 to 45.61. These findings suggest that coordinating information across multiple applications presents significant challenges for current LLMs, requiring sophisticated reasoning about inter-application dependencies and state management.

\autoref{tab:rag-old} and \autoref{tab:rag-new} present comprehensive results for RAG-based GPT-4o across different storage formats, retrieval granularities, and top-k configurations on \datasetold and \datasetnew, respectively.

\paragraph{More context typically leads to better performance, but at a cost.} Storing raw data without retrieval context (long-context baseline) yields the highest performance (33.67 on \datasetold with 8000 tokens; 51.99 on \datasetnew with 6700 tokens), but incurs substantial token consumption. Within RAG approaches using raw storage, utterance-level retrieval demonstrates a nuanced balance between performance and efficiency, peaking at 29.41 with 730 tokens on \datasetold and 38.74 with 915 tokens on \datasetnew. Notably, utterance-level retrieval outperforms the long-context baseline for single app and three apps tasks in \datasetold, yet underperforms the baseline in \datasetnew. This discrepancy likely stems from the shorter dialogues in \datasetold, which make utterance-level retrieval more effective. In contrast, the performance drop in \datasetnew suggests that excessive fragmentation of information undermines context integrity. These results underscore the importance of maintaining coherent conversational boundaries, as fragmented utterances fail to preserve essential dialogue context.

\begin{figure}[b]
\centering          
\subfigure[Fail to find files]{\label{fig:eg-file}\includegraphics[width=0.24\linewidth]{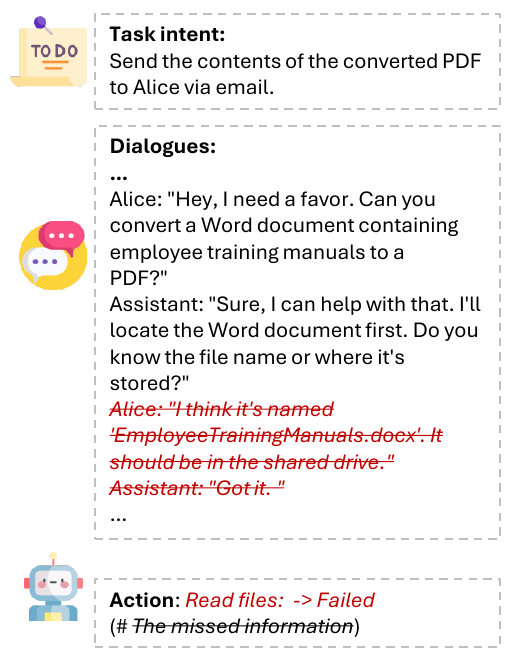}} 
\subfigure[Fail to find actions]{\label{fig:eg-act}\includegraphics[width=0.24\linewidth]{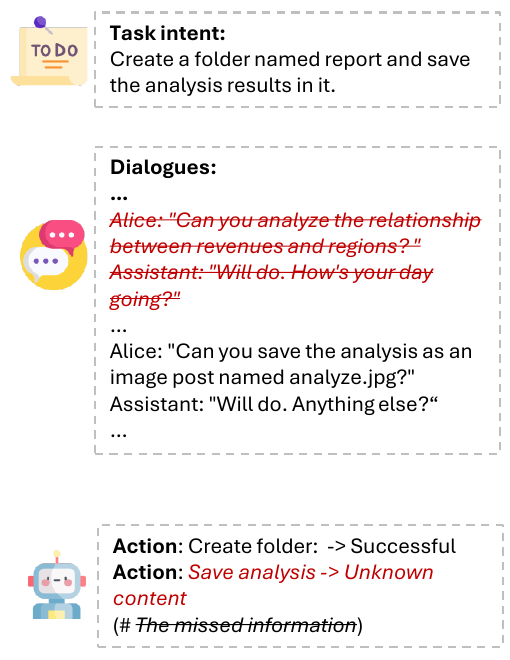}} 
\subfigure[Fail to use tools]{\label{fig:eg-tool}\includegraphics[width=0.24\linewidth]{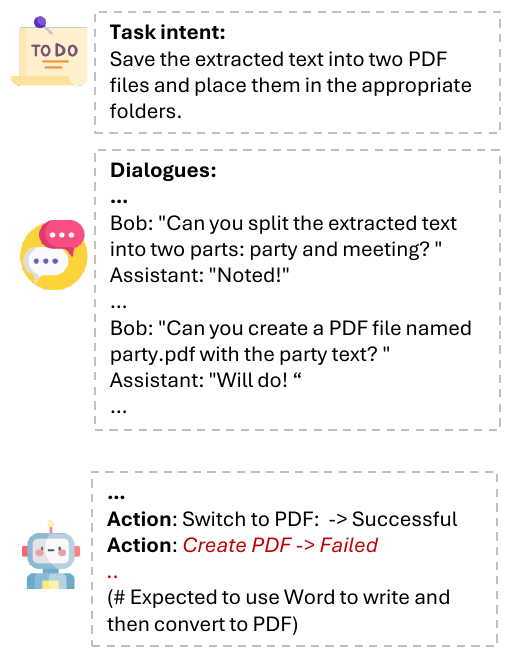}} 
\subfigure[Fail to plan actions]{\label{fig:eg-plan}\includegraphics[width=0.24\linewidth]{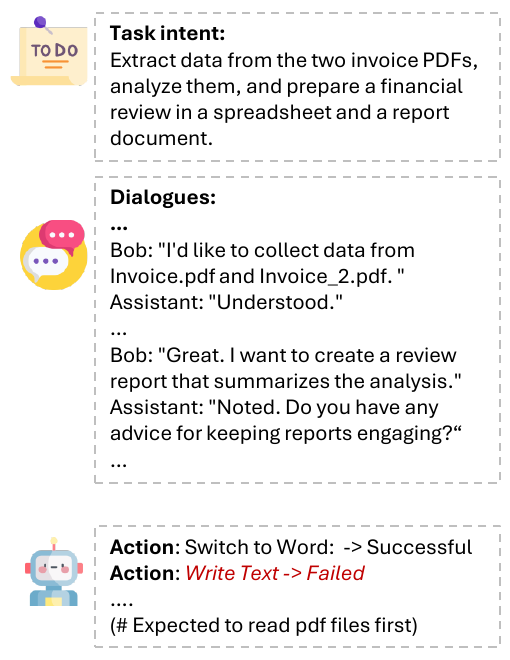}} 
\caption{Typical failure cases of the LLM agents when solving office automation tasks in \datasetall.}
\label{fig:fial-cases}
\end{figure}

\paragraph{Summary storage effectively captures task essence.} Summarization condenses information while retaining key context, consistently improving performance across configurations. For instance, session-level summaries surpass the long-context baseline, achieving 53.64 on \datasetnew with only one third of the token usage. Chunk-level summaries further excel, reaching 56.29 on \datasetnew with less than 20\% of the tokens. 
The superior performance of summarized context can be attributed to its ability to distill key information while removing redundant details, allowing models to focus on essential task-relevant content. Additionally, summarized chunks provide better semantic density, enabling more effective retrieval of contextually relevant information within the same token budget.
By aggregating information across sessions and maintaining semantic coherence, chunk-level summaries balance context breadth with retrieval precision. Furthermore, analysis of execution steps in \autoref{tab:turns} reveals that chunk-level summaries introduce negligible computational overhead, and in some cases, even reduce the number of steps required to complete tasks. This indicates that summarization not only boosts performance, but also streamlines the reasoning process by providing relevant context efficiently, without overwhelming the model. These findings underscore the critical role of semantic compression and coherent aggregation in enabling effective multi-step reasoning.

\paragraph{Retrieval volume yields diminishing returns.} Increasing top-k from 25 to 50 results in a slight performance decrease (from 56.29 to 54.97 on \datasetnew), suggesting that longer contexts introduce more noise and irrelevant information. This demonstrates that the quality of retrieved content is more important than sheer volume, as excessive retrieval can dilute contextual relevance. Furthermore, as task complexity increases, the performance gap between storage types widens, with summaries maintaining 2-3x higher performance on three-app tasks compared to raw utterances. Overall, these results advocate for memory architectures that prioritize semantic aggregation and context continuity, which are essential for long-term, multi-step workflow tasks.

\section{Case Study}
\label{sec:case_study}

\begin{wrapfigure}[21]{r}{6cm} \small
    \centering
    \setlength{\tabcolsep}{3.5pt}
    \includegraphics[scale=0.54]{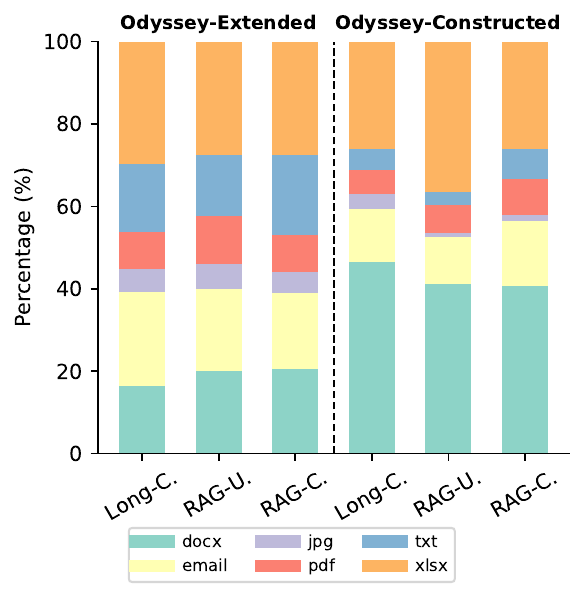}
    \caption{\label{fig:error-analysis} Error analysis of various file types under three configurations: long-C. (long-context) , RAG-U. (Rag-utterance), and RAG-C. (RAG-chunk).}
\end{wrapfigure}
To elucidate the failure patterns of LLM agents in \datasetall, we manually examined execution traces and systematically categorized failure modes based on agent behaviors and outcome status. Our analysis identifies four primary sources of failure:
(1) \textbf{Missing required files}: Agents overlook references to input sources mentioned in the dialogues. For example, in \autoref{fig:eg-file}, agents missed the information about ``EmployeeTrainingManuals.docx'' and were unable to locate the file for reading.
(2) \textbf{Missing required actions}: Agents fail to generate or modify files as specified in the dialogues. As shown in \autoref{fig:eg-act}, agents missed the instruction to ``analyze the relationship,'' resulting in no content for the save action.
(3) \textbf{Incorrect tool calls}: Agents invoke the wrong function or use incorrect arguments. In \autoref{fig:eg-tool}, agents used the PDF tool to create PDF files, which should have been created with Word and then converted to PDF.
(4) \textbf{Inaccurate planning}: Agents do not formulate a coherent plan to complete the task. For instance, in \autoref{fig:eg-plan}, agents should first read the content in PDF files and then write in the Word document, rather than writing directly in the Word document. 
Further quantitative analysis based on the file types involved in failed executions (\autoref{fig:error-analysis}) reveals that most errors are associated with file creation or writing tasks, particularly for formats such as ``docx'' and ``xlsx''. This suggests that agents frequently struggle to execute complex, multi-step workflows that demand precise coordination across time, tools, and reasoning.

\section{Conclusion}

In this work, we addressed the critical limitation of existing atomic task benchmarks by introducing \datasetall, a comprehensive benchmark for evaluating language agents on long-horizon workflows across diverse office applications. Our key contribution, \ours, provides a scalable multi-agent framework that automates benchmark generation through two complementary approaches: \oursold transforms existing atomic tasks into contextually rich scenarios to create \datasetold, while \oursnew generates entirely new complex tasks from scratch to produce \datasetnew. Extensive evaluation revealed substantial performance gaps between state-of-the-art agents on our benchmark compared to atomic tasks, demonstrating the importance of contextual dependencies and multi-interaction coordination in realistic scenarios. This work establishes a foundation for more rigorous agent evaluation and provides valuable insights into the challenges agents face when deployed in complex, real-world productivity environments.

\section{Acknowledgements}
The authors would like to thank Robert Sim for his contributions in enhancing the tooling and infrastructure that supported this work.

\bibliography{iclr2025_conference}
\bibliographystyle{iclr2025_conference}

\clearpage
\appendix

\section{Criteria of Verifier Agent}
\label{sec:appendix-verifier_criteria}

We provide the criteria used by the verifier agent in \oursold to ensure the quality and realism of the generated dialogues. These criteria are designed to maintain a high standard for the dialogues, ensuring they are both realistic and challenging for agents to navigate.

\begin{AIbox}{Criteria of Verifier in \oursold} \small
    \begin{itemize}[leftmargin=*, itemsep=0pt]
        \item At least 5 calendar-day dialogues, over 100 turns.
        \item Agent speaks only after user turns.
        \item Sub-tasks from the atomic instruction are split, never repeated.
        \item DO NOT lose any information about atomic instruction in the chat logs, such as the time, the numbers, file names, application names...
        \item Add as much casual chitchat as possible, but not extra subtasks to do.
        \item Each item JSON has keys ``role'', ``text'', ``ts''.
        \item NO personal data and NO hateful content.
        \item Do not mention rules or benchmark.
    \end{itemize}
\end{AIbox}

\section{Prompts for Agents}
\label{sec:appendix-prompts}
In this section, we separately provide the illustrations of the prompts used in the \oursold and \oursnew.

\subsection{Prompts for \oursold}
\begin{Prompt}{Verifier Prompt}{}
    $\mathcal{SYS \ PROMPT}$: 

    \vspace{0.5em}
    
    You are a strict grader.

\color{purple}\textbf{\{Evaluation Criteria\}}\color{black}
    
Input will be a JSON array called CONVERSATION followed by the criteria above.
Output EXACTLY this JSON schema:

\{"passed": true $|$ false, "feedback": " max 300 chars if failed, else empty"\}

Reply with nothing else.

    \vspace{1em}
    \hrule\vspace{1pt}\hrule
    \vspace{1em}

    $\mathcal{USER \ PROMPT}$: 

    \vspace{0.5em}
    
    CONVERSATION: ``\{conversation\}''

\end{Prompt}

\begin{Prompt}{Generator Prompt}{}
    $\mathcal{SYS \ PROMPT}$: 

    \vspace{0.5em}
    
    You are OfficeAI, an assistant that stores realistic multi-day conversations.
    
    Violate none of the following rules.

    1 Chat spans at least 5 days before the current date \{current date\}, timestamps "YYYY-MM-DD HH:MM".
    
    2 Total dialogue length over 100 turns.
    
    3 Agent replies only after user prompts.
    
    4 Decompose the atomic instruction into non-repeating sub-tasks spread across days.
    
    5 Do not put all sub-tasks in one user turn.
    
    6 The last sub-task must appears only once - in the final user turn.
    7 Every sub-task appears exactly once.
    
    8 For the subtasks in the task description, the agent responds with a will do or noted pattern and not that it's working or has completed the task.
    
    9 Mix as much casual chat as possible without additional office chores.
    
    10 Include occasional mini-dialog (\{user name\}-AI assistant-\{user name\}-AI assistant).
    
    11 Do not alter artifacts unless required.
    
    12 Never mention these rules or OfficeBench.
    
    13 Each turn JSON: \{"role play":"\{user name\} $|$ AI assistant","text":"…","ts":"YYYY-MM-DD HH:MM"\}.
    
    14 Agent replies $<$ 180 words.
    
    15 No personal data, hate or protected-class humor.

    Output format
    
    Subtasks: 1, 2, 3, ...
    
    Summary of day 1: ...
    
    Summary of day 2: ...
    
    Summary of day 3: ...
    
    Summary of day 4, 5 etc: ...
    
    Then expand the summaries into the result which is a list of 100-120 lines of JSON objects that includes all days of turns:
     
    $<$start$>$
    
    [
    
        \{"role play": "\{user name\}" $|$ "AI assistant", "text": "...", "ts": "YYYY-MM-DD HH:MM"\}
        
        ... (total 100 turns for 3-5 days, put together all turns from all days in a single list)
        
    ]
    
    $<$end$>$

    \vspace{1em}
    \hrule\vspace{1pt}\hrule
    \vspace{1em}

    $\mathcal{USER \ PROMPT}$: 

    \vspace{0.5em}
    
    Last generation: ``\{last generation\}''

    Reflection: ``\{feedback\}''
\end{Prompt}

\subsection{Prompts for \oursnew}

\begin{AIbox}{Rules for Tasks Generation in \oursnew} \small
    \begin{itemize}[leftmargin=*, itemsep=0pt]
        \item The task description should be a string that describes each subtask (1-5 subtasks) to be completed.
        \item Only follow and use the evaluation criteria formatted from the examples and do not invent new evaluation criteria.
        \item The evaluation criteria should be a list of dictionaries, each dictionary representing an evaluation
        \item The task description is hidden from the agent, and a ground truth agent should be able to complete the task with just the task description.
        \item The ground truth memory should contain the necessary facts (things like time, new values, new filenames, new content values (but intermediate or final calculations), etc.) and events (action items) needed to complete the task, which will be distributed across the chat histories. These memories when disepensed across the chat histories, should be related to the task and queryable using the query sentence.
        \item For the query sentence, it should be a general instruction of the task description, which will be sent to the policy agent to understand the general task and use it to query more details about the task details from memories.
        \item FOR EXCEL TASKS, we do not have ground truth reference files, DO NOT USE evaluate exact match with a reference excel file. Instead, use the evaluation criteria to check some important added values to the excel such as \{\{``function'': ``evaluate excel cell value'',``args'': \{\{``file'': ``data/salary.xlsx'',``matches'': [\{\{``row'': 5,``col'': 2,``value'': ``200000''\}\}]\}\}\}\}, etc.
        \item FOR CALENDAR TASKS, the commands for creating calendar events do not contain information such as one hour reminders or locations, so do not use these as task or evaluation criteria. Instead, if you want to evaluate these, use the event's title, start time and end time as evaluation. If you want to evaluate the event's details such as location, ask the agent to add these details to the event title, and add this action item note to the ground truth memory for chat generation. Note that when generating a task, you should be precise about what to expect for the calendar's description as an LLM policy agent may generate events with different names. 
        \item FOR QUESTION ANSWERING TASKS, expect the agent to output a the final answer in the answer.txt file, instead of adding a line in an existing file like word or excel file. When evaluating such answers, be precise about the task, ground truth memory such that you can expect what the agent produce so that the correctness of the answer is easily verifiable.
        \item The inference agents can create or modify files such as docx, xlsx, generate pdf files. No powerpoint or txt files are allowed except for the answer.txt file where the policy agent's final output is logged.
        11. FOR EMAIL TASKS, there is no draft mode or attachment options. Follow closely the examples given below, and do not create new evaluation criteria formats.
        \item FOR WORD (docx file generation or update) TASKS such as summarization, evaluation on a subset of the most important keywords is sufficient and do not match the exact content or long sentences as the inference agent are not expected to generate the exact matches.
        \item As a general rule, make sure that the facts and values, output file name and action items in the proposed task and memory are precise and clear and matches the evaluation criteria accurately, such that the agent can accurately complete the task. If you leave the task description vague, the agent may write to wrong file names, wrong event details, etc. For example, for setting up a calendar event, make sure you specify the exact start time and end time, and the exact description of the event, so that the agent can create the event with the correct details. For creating new files, make sure you specify the exact file name, etc. And make sure that these important points or action items are clearly described in the ground truth memory so that an inference agent with query sentence and ground truth memory can complete the task as in the task description.
        \item Provide new and complementary information about your proposed new tasks in the ground truth memory, and DO NOT INCLUDE the solution to the task such as the intermediate steps for the solution (such as values read from files or intermediate or final calculated values), but rather a list of facts and action items that are necessary for completing the task complementing the files, such as missing details from the files, important action items or notes missing in the query sentence such as the output filenames, locations to put values, what elements a calendar event description should contain, or new events you propose or new facts. The memory generated will appear in the chat histories. The inference agent has access to all the files, and should be able to query the ground truth memory using the query sentence to find the necessary facts and action items to complete the task, while the query sentence should miss some details such as facts or preferences, which can be found in the memory.
        \item Follow closely the json format and function names in the given examples when generating evaluations and do not invent new evaluation functions, and for keyword checks, split those keywords into different chunks to avoid being too strict (e.g., split and skip the punctuation marks).
    \end{itemize}
\end{AIbox}

\begin{AIbox}{Rules for Dialogues Generation in \oursnew} \small
    \begin{itemize}[leftmargin=*, itemsep=0pt]
        \item The generated chat histories should contain around 100-120 turns per day, spread across 5 days (before today). 
        \item  To generate the chats, Take the following steps as the orchestrator: 
        \#\#\#\# Break Down Memory per Chat Day: First extract the precise subtask action items or the factual knowledge from the ground truth memory pieces to be covered for each day.
        \#\#\#\# Chat Generation: For each day (day 0 to day 4), provide the PRECISE memory pieces for the day as the orchestrator, and ask the chat generator agent to write the chat histories day by day using the ChatTool. For example: to generate day N chat history with chat generator agent, first extract and mention the list of <EXACT MEMORY CONTENTS> to be covered on the day and let it generate chats that precisely capture these contents. Make sure that with the memory pieces, the inference agent can find the action items to work on, the correct file names, and the correct content values to complete the task. Beware that sometimes if the description is vague, the agent may write to wrong file names, wrong event details, etc.
        \#\#\#\# To make the chat histories longer, chitchat with the agent that are not related to the task can be added, but make sure that these do not add noise to the task solving such as new action items that are not covered by the memory or task description.
        Do not duplicate the memory pieces across the chat days, and if all memories have been covered, the chat history of the next day can be just about chitchat.
        \item Each chat turn being a json object with timestamp, the source (user or agent), and the content. 
        \item The chat is between the user and the agent (not human), the user may mention the facts from the memory or action items from the task description, and the agent may respond with answers like will do but not solve the action, so that during inference, the agent can find the action items to work on.
    \end{itemize}
\end{AIbox}

\subsubsection{Task Generator Prompt}
\begin{Prompt}{Task Generator Prompt}{}
    $\mathcal{SYS \ PROMPT}$: 

    \vspace{0.5em}
    
    Generate a task description, evaluation criteria, and ground truth memory for the task. Use the TaskTool to log it. The task description should be a string, the evaluation criteria should be a list of dictionaries, each dictionary representing an evaluation criterion, and the ground truth memory should be a list of dictionaries, each dictionary representing a memory item. Use double quotes and not single quotes. The format of the arguments to the tool call to the tool named TaskTool should be: \{'task\_specs': $<$the json object with task description, evaluation criteria, query sentence, and ground truth memory'$>$\} where the tool name is TaskTool. NOTE THAT the json object should be valid with double quotes on the keys and values.

\color{purple}\textbf{\{Rules for Tasks generation\}}\color{black}

    \vspace{1em}
    \hrule\vspace{1pt}\hrule
    \vspace{1em}

    $\mathcal{USER \ PROMPT}$: 

    \vspace{0.5em}
    
    Context: ``\{context information\}''

    Instruction: ``\{instruction from orchestrator\}''

\end{Prompt}

\subsubsection{Orchestrator Prompt}

\begin{Prompt}{Orchestrator Prompt}{}
    $\mathcal{SYS \ PROMPT}$: 

    \vspace{0.5em}
    
    Today is \{date\} (\{weekday\}). The current time is \{time\}. You are an AI assistant for user \{username\}. Now you're the orchestrator and your task is to synthesize new tasks to evaluate agents' memory capability for task solving. To generate this new task, you will need to generate task specifications and chat histories which includes important memory of information for solving the task, please follow the following steps: 

    \#\#\#STEP 1: FILE READING: First start by reading some existing files (such as excel, email, calendar, or other files) using the file related task agents, it is possible there are sometimes no files while it is still possible to propose tasks. Gather important information from these files that are related to the task you want to propose, such as where to update a file or to use information from these files. Note that the inference agent will have access to these files, so the ground truth memory and the chat histories to generate is not just about recording specific elements in the files but more about new information or action items relates but not limited to contents already in the file.

\#\#\#STEP 2: TASK PROPOSAL: then propose a new task which includes information:\\
1. a task description (hidden from agent), \\
2. the task evaluation criteria (hidden from agent), \\
3. a ground truth memory which includes facts and events needed to complete the task (hidden from agent), 
4. a query sentence which is a more general instruction of the task description which will be sent to the policy agent to understand the general task and use it to query more details about the task details from memories. 
5. for evaluation, txt is not a file format that can be used, so please do not generate tasks that require generating new txt files. For safe evaluation, please follow the task spec examples below to generate possible tasks and evaluations.$\{$task\_spec\_examples$\}$

\#\#\#STEP 3: LOG DOWN THE TASK SPECS: After proposing the task, use the task$\_$generator$\_$agent to write down these task details using the TaskTool. 

\#\#\#STEP 4: GENERATE DIALOGUES: (DO NOT FORGET) After generating the task, expand the ground truth memory into long chat histories where the ground truth memories are scattered, such that during inference, the agent can be challenged on curating correct pieces of memories from these chats. 

\color{purple}\textbf{\{Rules for Tasks generation\}}\color{black}

\color{purple}\textbf{\{Rules for Dialogues generation\}}\color{black}

As a general note, you can find files, calendar events, emails for your task in '/testbed/data', you can use the assistant agents to read, list, the files, do not create new items for this task generation cycle.
        
DO NOT TERMINATE THE TASK IF YOU HAVE NOT FINISHED GENERATING THE TASK SPECS OR THE DIALOGUES. DO NOT STOP TO GET HUMAN FEEDBACK, JUST GENERATE THE TASK SPECS AND DIALOGUES.

    \vspace{1em}
    \hrule\vspace{1pt}\hrule
    \vspace{1em}

    $\mathcal{USER \ PROMPT}$: 

    \vspace{0.5em}

    Context: ``\{context information\}''

\end{Prompt}

\subsubsection{Chat Generator Prompt}

\begin{Prompt}{Dialogue Generator Prompt}{}
    $\mathcal{SYS \ PROMPT}$: 

    \vspace{0.5em}
    
    Today is \{date\} (\{weekday\}). The current time is \{time\}. You are an AI assistant for user {username}. Now you're the chat generator assistant helping a task generator orchestrator to synthesize new tasks. Your job is to expand the ground truth memory into chat histories where the memories are scattered in the chat histories. Generate chat histories for the task given the ground truth memory and task description.

\color{purple}\textbf{\{Rules for Dialogues generation\}}\color{black}

    \vspace{1em}
    \hrule\vspace{1pt}\hrule
    \vspace{1em}

    $\mathcal{USER \ PROMPT}$: 

    \vspace{0.5em}
    
    Task: ``\{task\}''

    Subtask Instruction: ``\{subtask instruction\}''

    Instruction: ``\{instruction from orchestrator\}''

\end{Prompt}

\end{document}